\documentclass[10pt,twocolumn,letterpaper]{article}

\usepackage{cvpr}              % To produce the CAMERA-READY version
\usepackage[dvipsnames]{xcolor}

\definecolor{cvprblue}{rgb}{0.21,0.49,0.74}
\usepackage[pagebackref,breaklinks,colorlinks,citecolor=cvprblue]{hyperref}

\usepackage{graphicx} % for pdf, bitmapped graphics files
\usepackage{amsmath} % assumes amsmath package installed
\usepackage{amssymb}  % assumes amsmath package installed
\usepackage{bm}
\usepackage{cite}
\usepackage{multirow}
\usepackage{array}
\usepackage{makecell}
\usepackage{threeparttable}

\def\paperID{9699} % *** Enter the Paper ID here
\def\confName{CVPR}
\def\confYear{2024}

\title{Adaptive VIO: Deep Visual-Inertial Odometry with Online Continual Learning}

\author
{Youqi Pan\qquad Wugen Zhou\qquad Yingdian Cao\qquad Hongbin Zha\\
National Key Lab of GAI,\quad Institute for AI,\quad School of IST\\
PKU-SenseTime Joint Lab of MV\\
Peking University\\
{\tt\small panyouqi@stu.pku.edu.cn\quad \{zhouwugen, yingdianc\}@pku.edu.cn\quad zha@cis.pku.edu.cn}
}

\begin{document}
\maketitle
\begin{abstract}
Visual-inertial odometry (VIO) has demonstrated remarkable success due to its low-cost and complementary sensors. However, existing VIO methods lack the generalization ability to adjust to different environments and sensor attributes. In this paper, we propose \textbf{Adaptive VIO}, a new monocular visual-inertial odometry that combines online continual learning with traditional nonlinear optimization. Adaptive VIO comprises two networks to predict visual correspondence and IMU bias. Unlike end-to-end approaches that use networks to fuse the features from two modalities (camera and IMU) and predict poses directly, we combine neural networks with visual-inertial bundle adjustment in our VIO system. The optimized estimates will be fed back to the visual and IMU bias networks, refining the networks in a self-supervised manner. Such a learning-optimization-combined framework and feedback mechanism enable the system to perform online continual learning. Experiments demonstrate that our Adaptive VIO manifests adaptive capability on EuRoC and TUM-VI datasets. The overall performance exceeds the currently known learning-based VIO methods and is comparable to the state-of-the-art optimization-based methods.
\end{abstract}

\section{Introduction}
\label{sec:intro}
Obtaining reliable motion estimation in unknown environments is critical to many vision and robotics tasks, such as augmented reality (AR), unmanned aerial vehicle (UAV), and autonomous driving. Simultaneous localization and mapping (SLAM) is one of the critical approaches that employs onboard sensors to estimate agent trajectory and build a map of local environments. Researchers have extensively investigated visual-inertial SLAM (VI-SLAM) and visual-inertial odometry (VIO) due to their low-cost and complementary sensors. It often presents a more accurate and robust trajectory estimation than visual odometry (VO) or inertial odometry (IO).

% different frameworks 
\begin{figure}[t]
  \centering
  \includegraphics[width=0.48\textwidth]{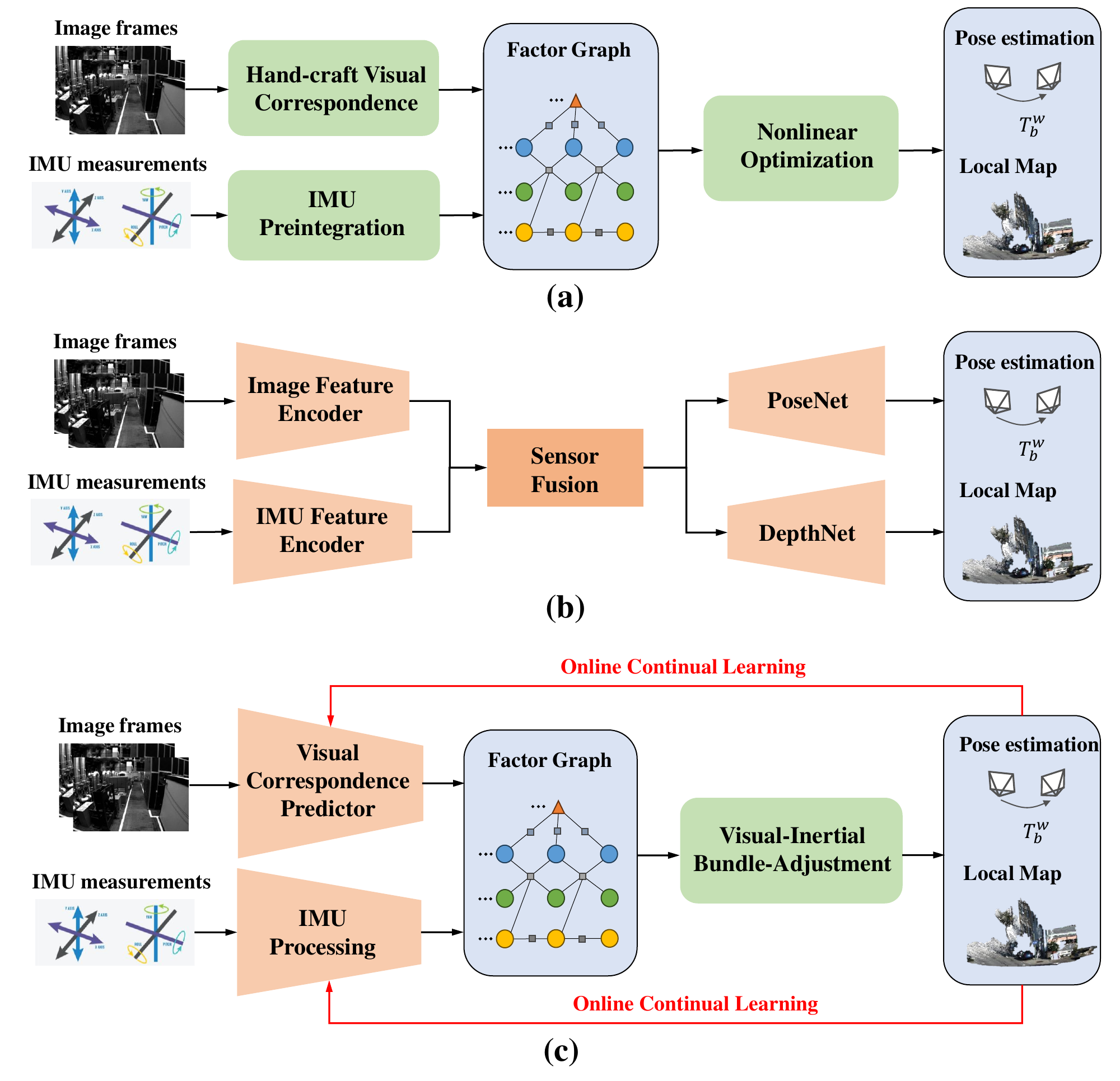}
  \caption{Frameworks of different VIO methods. Learning-based modules are colored in orange. Traditional computational modules are colored in green. \textbf{(a)} Classic optimization-based method. \textbf{(b)} End-to-end learning-based method. \textbf{(c)} Learning-optimization-combined method with online continual learning (ours).}
  \label{pipline compare}
\end{figure}

% Classic VIO systems can be divided into filter-based methods, like MSCKF, and optimization-based methods, such as VINS-mono, ORB-SLAM3, and DM-VIO. 
A classic VIO system is mainly composed of visual association, IMU preintegration, and back-end nonlinear optimization \cite{okvis, vins, vidso, wangxin, dmvio} (or filtering \cite{msckf2, okvis}), as shown in Fig.\ref{pipline compare}(a). Classic methods are featured in the systematic framework and fine-grained pipeline, working well in favorable conditions. However, they are less accurate and may even fail in challenging scenarios (\emph{e.g.} low-light condition, abrupt movement), which can be attributed to the reliance on low-level and hand-crafted visual features. In addition, trajectory drift caused by IMU bias is also one of the critical reasons affecting the system's performance, while traditionally modeling IMU bias as a random walk is insufficient to reflect its evolutionary patterns.

In order to overcome the shortcomings of classic methods that rely on pre-defined features, many learning-based approaches have been proposed. End-to-end learning-based VO can extract features from image streams and directly generate pose and depth estimation without explicit optimization, which has shown promising results in recent years. Similarly, some end-to-end learning-based VIO methods are developed. As shown in Fig.\ref{pipline compare}(b), these methods typically use two separate networks to extract image and IMU features, fuse them through a fusion network, and subsequently get pose and depth estimations from networks. Nonetheless, these methods suffer from poor generalization in complex motion scenarios, with lower accuracy than classic methods.

In this paper, we propose a novel VIO system named Adaptive VIO. As shown in Fig.\ref{pipline compare}(c), unlike classic methods or end-to-end learning-based VIO, our approach integrates learning with classic computations. We leverage the strengths of neural networks in predicting visual correspondence and IMU bias, replacing the traditional optical flow or hand-craft feature matching and random walk modeling for IMU bias. On the other hand, the refined estimates obtained through classic optimization serve as feedback information, generating loss functions for the predictor network, thus enabling self-supervised learning. Finally, thanks to the learning-optimization-combined framework and feedback mechanism, we can conduct online continual learning, enabling the VIO system to adapt across different environments and achieve better tracking performance.

Our main contributions can be summarized as follows:
\begin{itemize}
    \item We propose a novel learning-optimization-combined VIO, which predicts visual correspondence and IMU bias through learning approaches and obtains accurate state estimation through classic nonlinear optimization.
    % \item We introduce a feedback mechanism for the visual-inertial odometry, utilizing pose and depth estimates from nonlinear optimization to construct loss functions, which provide feedback for updating both the visual correspondence network and the IMU bias estimation network through self-supervised learning.
    \item We introduce a feedback mechanism for the system, utilizing the estimation from nonlinear optimization to construct loss functions, updating the networks in a self-supervised manner.
    
    \item We develop online continual learning to refine the networks in different scenarios. Experiments demonstrate the strong generalization and adaptive capabilities of our method, yielding overall results comparable to state-of-the-art VIO systems.

\end{itemize}

% our pipeline
\definecolor{darkyellow}{HTML}{B8860B}
\begin{figure*}
  \centering
  \includegraphics[width=1.0\linewidth]{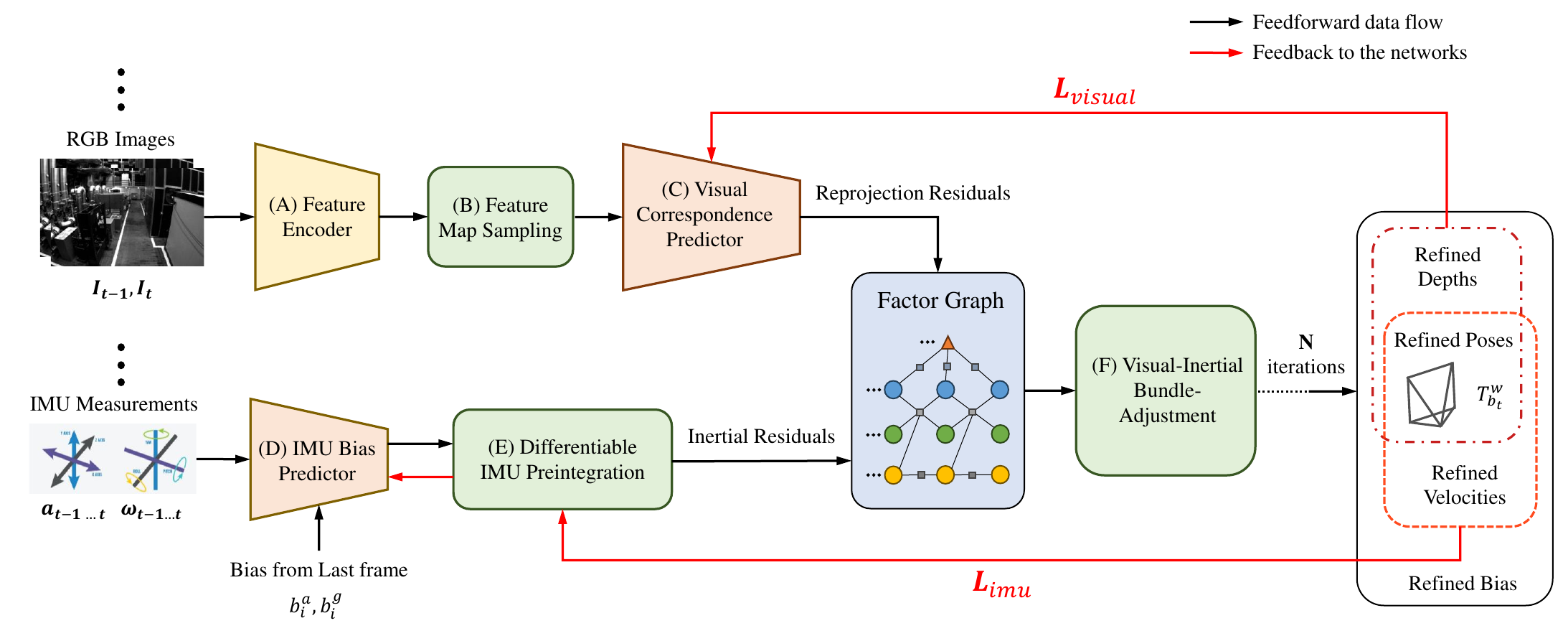}
  \caption{The tracking pipeline of our VIO. The green modules (B, E, F) denote manually designed algorithms. The yellow (A) and orange (C, D) trapezoids represent modules implemented by neural networks, and orange modules can get online continual learning.}
  \label{pipline}
\end{figure*}

\section{Related Work}
\label{sec: rlworks}

\subsection{Classic VIO}
In the past decade, VIO has been an active topic in the field of SLAM. Due to the complementary sensors, VIO exhibits enhanced robustness compared to VO across various scenarios and makes scale observable in monocular setups.

Classic VIO systems are mainly based on tightly coupled approaches, wherein visual and IMU constraints are fused through filtering or nonlinear optimization. Filter-based methods, such as MSCKF \cite{msckf, msckf2}, and ROVIO \cite{rovio}, use the extended Kalman filter (EKF) to propagate and update the current state. While nonlinear-optimization-based methods, like VINS \cite{vins, vinsfusion}, ORB-SLAM3 \cite{orbslam3} and DM-VIO \cite{dmvio}, adopt local visual-inertial bundle adjustment for the state estimation, achieving more accurate pose tracking. 

Classic VIO uses optical flow or hand-craft features for visual association, constructing motion constraints based on photometric or reprojection errors. As for IMU processing, modeling IMU bias as the random walk is a common practice \cite{manifold}. Classic VIO methods have gained widespread application, but the manually defined data association and IMU bias modeling are often inaccurate in challenging scenarios, leading to suboptimal results or even failure.

\subsection{Learning-based VO and VIO}
In recent years, learning-based methods in VO, IO, and VIO have been extensively researched\cite{sfmlearners, russell, selfvio}, yielding promising results. 

End-to-end learning-based VO utilizes pose and depth estimation networks to replace classic modules of tracking and mapping, trained in either a supervised \cite{deepvo, xue} or self-supervised \cite{sfmlearners, li-adver} manner. Some end-to-end learning-based VIO approaches have also been proposed \cite{vinet, violearner, deepvio}. For instance, SelfVIO \cite{selfvio} leverages networks to encode and adaptively fuse visual and IMU information, estimating depth and pose by self-supervised learning as VO.

Despite promising results on some datasets, end-to-end learning-based VO and VIO exhibit lower accuracy than classic approaches, particularly struggling to get accurate pose estimates under complex motions. Therefore, some propose to combine learning techniques with traditional modules \cite{banet, codeslam, rninvio}. DROID-SLAM \cite{droid} and DPVO \cite{dpvo} combine iterative visual correspondence updates with differentiable bundle adjustment, demonstrating excellent performance across multiple datasets. iSLAM \cite{islam} combines learning-based VO with graph optimization, further enhancing performance through self-supervised learning. From the IMU perspective, learning-based data preprocessing and bias estimation techniques are also proposed \cite{russell, tlio, rninvio, imuprocessing}. Zhang et al. \cite{imuprocessing} learn to denoise IMU measurements and use preintegration loss derived from ground truth poses to train the network. Buchanan et al. \cite{russell} explicitly propose models for the dynamics of bias by networks and incorporate them into factor graph optimization, replacing traditional random walk models.

These methods combining learning with classic computations provide us with significant inspiration. To the best of our knowledge, our approach is the first to simultaneously integrate learning-based visual association and IMU bias modeling into a VIO framework and refine them by self-supervised learning.

\subsection{Online continual learning}

Generally speaking, classic VO/VIO algorithms are manually pre-defined, and learning-based ones are usually pre-trained on dedicated datasets and then inference on scenes with similar distribution. These methods may suffer from domain shift problems if directly applied to a different scenario, leading to inferior performance. 

To address this issue, Li et al. \cite{li-meta, li-online} propose an online-learning VO framework to generalize better on unseen environments and use meta-learning techniques to facilitate fast adaptation. V{\"o}disch et al. \cite{clslams, covio} introduce online continual learning for SLAM, incorporating a replay buffer to prevent catastrophic forgetting and using asynchronous network updates to optimize system efficiency.

While these online learning methods have shown adaptability in autonomous driving scenarios such as Cityscape\cite{cityscapes} and KITTI \cite{kitti}, they often fail in complex environments and under complex motions. In our method, the states from optimization provide feedback signals to the networks, enabling self-supervised online continual learning, which has proved effective in EuRoC \cite{euroc} and VI-TUM datasets \cite{tumvi}.

\section{Method}
\label{sec: method}

In this section, we introduce our monocular VIO method in detail. We start with the unique framework design, which combines classic optimization with deep learning techniques (Sec.\ref{subsec: framework}). Next, we present the feedback mechanism and self-supervised updates for the networks (Sec.\ref{subsec: network}). Finally, we delve into more details of the VIO system, focusing on online continual learning, which is pivotal for the system to adapt to diverse scenarios and achieve improved performance gradually (Sec.\ref{subsec: online}).

\subsection{Learning-optimization-combined framework}
\label{subsec: framework}
The tracking pipeline of our VIO system is illustrated in Fig.\ref{pipline}, featured by the framework that integrates learning and classic optimization. The orange shapes in the diagram represent computations performed by the neural network, while the blue shapes indicate traditional manual calculations.

\textbf{(A) Feature encoder} receives the latest RGB image frame and encodes it into feature maps by convolutional neural networks, providing stable feature encoding for subsequent visual correspondence.

\textbf{(B) Feature map sampling} module randomly selects several feature points and their neighborhood from the feature map, generating feature patches. These patches are considered as keypoints, facilitating subsequent matching and depth and pose estimation. The module can be differentiable{ \cite{dpvo}}, which made the gradient backpropagation possible. 

\textbf{(C) Visual correspondence predictor} predicts visual matching relationship among the keyframes in the factor graph. It takes initial matches generated by reprojection as input and outputs the updates relative to the initial matching, which can also be viewed as reprojection residuals. The pose and depth estimation are generated from IMU state propagation or visual-inertial bundle adjustment, which will be detailed in the following content.

\textbf{(D) IMU bias predictor} takes IMU bias estimation from the previous time step, along with the accelerometer and gyroscope measurements between previous and current image frames, as inputs to predict the IMU bias  $\hat{\mathbf{b}}_a$, $\hat{\mathbf{b}}_g$ for the current time step. The IMU timestamps and bias configuration follow the settings in \cite{vins, orbslam3}. Specifically, we synchronize the IMU preintegration interval with the timestamps of the image frames, assuming that the bias remains fixed within the interval.

\textbf{(E) Differentiable IMU preintegration} integrate IMU data in body poses of the previous image frame, which is independent of the initial conditions and can be treated as a single observation between two adjacent image frames. Assuming the timestamps of $i^{th}$ and ${i+1}^{th}$ IMU frames are between the timestamps of $k^{th}$ and ${k+1}^{th}$ image frames. Let $\hat{\bm{\alpha}}_{i}^{b_k}$, $\hat{\bm{\beta}}_{i}^{b_k}$, $\hat{\bm{\gamma}}_{i}^{b_k}$ represent the preintegration of translation, velocity and rotation until the $i^{th}$ IMU frame, where $b_k$ denotes the preintegration is under the body coordinate at the $k^{th}$ image frame. The body coordinate aligns with the IMU coordinate. Then, the preintegration until ${i+1}^{th}$ IMU frame can be expressed in the following form\cite{vins}:

\begin{equation}
    \hat{\bm{\alpha}}_{i+1}^{b_k} = \hat{\bm{\alpha}}_{i}^{b_k} + \hat{\bm{\beta}}_{i}^{b_k}{\delta}t + \frac{1}{2}\mathbf{R}(\hat{\bm{\gamma}}_{i}^{b_k})\hat{\mathbf{a}}_{i}{\delta}t^2 
\end{equation}
\begin{equation}
		 \hat{\bm{\beta}}_{i+1}^{b_k} = \hat{\bm{\beta}}_{i}^{b_k} + \mathbf{R}(\hat{\bm{\gamma}}_{i}^{b_k})\hat{\mathbf{a}}_{i}{\delta}t
\end{equation}
\begin{equation}
		 \hat{\bm{\gamma}}_{i+1}^{b_k} = \hat{\bm{\gamma}}_{i}^{b_k} \otimes \begin{bmatrix}
			\frac{1}{2}\hat{\bm{\omega}}_{i}{\delta}t \\ 1
		\end{bmatrix}
\end{equation}
where $\hat{\mathbf{a}}_{i}$, $\hat{\bm{\omega}}_{i}$ are acceleration and angular velocity measurements of $i^{th}$ IMU frame, after subtracting the network-predicted bias $\hat{\mathbf{b}}_a$, $\hat{\mathbf{b}}_g$, respectively. $\otimes$ represents the product in quaternion. 

The propagation from $i^{th}$ to ${i+1}^{th}$ IMU frame is differentiable with respect to $\hat{\mathbf{a}}_{i}$ and $\hat{\bm{\omega}}_{i}$. Henceforth, the preintegration from the ${k^{th}}$ to ${k+1}^{th}$ image frame is also differentiable, which allows the gradient backpropagate from preintegration to the IMU bias predictor.

The propagation of covariance and approximate bias update techniques are also adopted in our system, which are detailed in \cite{manifold}. The differentiation and backpropagation of Lie Group can refer to \cite{sola2018micro, lietorch, pypose}.

\textbf{(F) Visual-inertial bundle adjustment (VIBA)} are standard techniques for solving state variables in classic VIO systems, which we also adopt with some distinctions.

Our factor graph comprises states within a sliding window of $n$ image frames, incorporating visual constraints, IMU constraints, and their interrelations. The full state vector in the factor graph is:
\begin{equation}
        \mathcal{X} = \{ {\mathbf{p}^{w}_{b}}, \mathbf{q}^{w}_{b},     {\mathbf{v}^{w}_{b}}, {\mathbf{b}_{a}}, {\mathbf{b}_{g}}, {\mathbf{d}_{0..m}}\}_{0..n}
\end{equation}
where $\mathbf{p}^{w}_{b}$, $\mathbf{q}^{w}_{b}$ and $\mathbf{v}^{w}_{b}$ denote translation, quaternion and velocity of the body. Bolded $\mathbf{b}_a$ and $\mathbf{b}_g$ are biases of the accelerator and gyroscope. ${\mathbf{d}_{0..m}}$ represents the depth of $m$ features from the frame.

The visual constraints in the factor graph are reprojection residuals, which reflect the coordinate error between the matches predicted by the network and the matches projected under current states. We construct reprojection residuals for each feature, and the network provides the confidence of each residual.

\textit{Reprojection residuals}:
\begin{equation}
    \begin{split}
        \mathbf{r}_v = \sum_{i,j\in\mathcal{G}}\|\hat{p}_j^t - KT^t_s\mathbf{d}(p_i^s)K^{-1}p_i^s\|
    \end{split}
\end{equation}
where $p_i^s$, $\hat{p}_j^t$ are a pair of matching points predicted by the network. $T^t_s$ denotes the coordinate transformation between the frames of the matching points in the image. $K$ is intrinsic to the camera. All such matching points constitute visual constraints in the factor graph $\mathcal{G}$.

IMU constraints can be categorized into two types: preintegration residuals and bias residuals. For two consecutive image frames $I_{k}$ and $I_{k+1}$, the following expressions are given:
\begin{equation}
    \mathbf{r}_{\mathcal{I}_{k+1}^{k}} = \{\mathbf{r}_{\Delta \mathrm{p}_{k+1}^{k}}, \mathbf{r}_{\Delta \mathrm{v}_{k+1}^{k}}, \mathbf{r}_{\Delta \mathrm{R}_{k+1}^{k}}, \mathbf{r}_{\Delta \mathrm{(b_a)}_{k+1}^{k}}, \mathbf{r}_{\Delta \mathrm{(b_g)}_{k+1}^{k}}\}
\end{equation}

\textit{Preintegration residuals:}
\begin{equation}
    \begin{split}
        \mathbf{r}_{\Delta \mathrm{p}_{k+1}^{k}} = &\mathrm{R}_{w}^{b_k} \Big( {\mathbf{p}}_{b_{k+1}}^{w} - {\mathbf{p}}_{b_{k}}^{w} - {\mathbf{v}}_{b_{k}}^{w} \Delta t_k  \\& + \frac{1}{2} g^{w}\Delta t_k^2 \Big) - \hat{\bm{\alpha}}^{b_k}_{b_{k+1}}(\mathbf{b}_a,\mathbf{b}_g)
    \end{split}
    \label{eq:ppre}
\end{equation}
\begin{equation}
    \begin{split}
        \mathbf{r}_{\Delta \mathrm{v}_{k+1}^{k}} = &\mathrm{R}_{w}^{b_k} \Big( {\mathbf{v}}_{b_{k+1}}^{w} - {\mathbf{v}}_{b_{k}}^{w} +  g^{w}\Delta t_k \Big) \\& - \hat{\bm{\beta}}^{b_k}_{b_{k+1}}(\mathbf{b}_a,\mathbf{b}_g)
    \end{split}
    \label{eq:vpre}
\end{equation}
\begin{equation}
    \begin{split}
        \mathbf{r}_{\Delta \mathrm{R}_{k+1}^{k}} = &\mathrm{Log} \left( \mathbf{R}(\hat{\bm{\gamma}}_{b_{k+1}}^{b_k}(\mathbf{b}_g))^\mathrm{T} \mathrm{R}_{w}^{b_k} \mathrm{R}_{b_{k+1}}^{w} \right)
    \end{split}
    \label{eq:rpre}
\end{equation}
where $\hat{\bm{\alpha}}^{b_k}_{b_{k+1}}$, $\hat{\bm{\beta}}^{b_k}_{b_{k+1}}$, $\hat{\bm{\gamma}}_{b_{k+1}}^{b_k}$ denotes preintegration terms and $g^w$ is gravity under the world coordinate.

\textit{Bias residuals}:
\begin{equation}
    \begin{split}
        \mathbf{r}_{\Delta \mathrm{(b_a)}_{k+1}^{k}} = (\mathbf{b}_a)^k - (\hat{\mathbf{b}}_a)^k
    \end{split}
    \label{eq:ba}
\end{equation}
\begin{equation}
    \begin{split}
        \mathbf{r}_{\Delta \mathrm{(b_g)}_{k+1}^{k}} = (\mathbf{b}_g)^k - (\hat{\mathbf{b}}_g)^k
    \end{split}
    \label{eq:bg}
\end{equation}
where $\hat{\mathbf{b}}_a$, $\hat{\mathbf{b}}_g$ are the bias predicted by the network.

We model accelerator and gyroscope bias as a Gaussian distribution with the network-predicted bias value as the mean, and the variance is manually set. Similar methods are also elaborated in \cite{russell}.
In addition to the differences in bias residuals compared to traditional random walk methods, there are other distinctions in treating bias. Firstly, we only conduct preintegration by the network-predicted bias ($\hat{\mathbf{b}}_a$, $\hat{\mathbf{b}}_g$). We never reintegrate measurements by the updated bias (${\mathbf{b}}_a$, ${\mathbf{b}}_g$) after optimization. Second, the updated bias from the current frame serves as inputs to the predictor, along with measurements from the next frame, generating a new bias prediction instead of being directly passed to the next time step.

% Compared to classic methods, our approach leverages neural networks to uncover the inherent dynamics of bias. Our strategy saves computational costs and prevents erroneous bias estimates from propagating downward, enhancing system performance and stability.

The final optimization objective is composed of IMU residuals and visual residuals, which can be written as: 
\begin{equation}
    \begin{split}
        \mathcal{X}^* = \underset{\mathcal{X}} {\arg\min} \Big( \|\mathbf{r}_\mathrm{v}\|_{\Sigma_p}^2 + \sum_{k=0}^n \| \mathbf{r}_{\mathcal{I}_{k+1}^{k}} \|_{\Sigma_{\mathcal{I}_{k+1}^{k}}}^2 \Big)
    \end{split}
\end{equation}

Considering the Gauss-Newton method is naturally differentiable, we use it to solve the factor graph, iterating twice for each timestamp. This facilitates gradient backpropagation to the neural network during training and online adaptation.

\subsection{Feedback and self-supervised update}
\label{subsec: network}
In our approach, factor graph optimization can provide feedback signals for visual and IMU networks, enabling self-supervised learning updates.

The feature encoder and visual correspondence predictor adopt similar network structures to DPVO \cite{dpvo}. The difference is that our predictor does not compose a hidden state, making the inference of the correspondences in each iteration an independent process irrelevant to the previous network state. The modifications make our pipeline more akin to the classic SLAM, enhancing system interpretability.

The poses and depths optimized by VIBA can be used to construct photometric loss through reprojection:

\begin{equation}
    \begin{split}
        \mathbf{L}_{visual} = \sum_{i\in\mathcal{G}}\|I^t({p}_i^*) - I^s(p_i)\|
    \end{split}
    \label{eq:pholoss}
\end{equation}
\begin{equation}
    \begin{split}
        {p}_i^* =  KT^t_s\mathbf{d}(p_i^s)K^{-1}p_i^s
    \end{split}
\end{equation}
where $I$  represents the intensity of the pixel.

The bias prediction network consists of normalization layers, fully connected layers, and a GRU \cite{gru} module. 
The bias updated from the previous timestamp is initially normalized and then encoded as hidden states for the GRU. Simultaneously, the IMU measurements are normalized and concatenated with the normalized bias. After encoding by a fully connected layer, they serve as inputs to the GRU. The GRU's output is then transformed into the current bias estimation via another fully connected layer.

After visual-inertial bundle adjustment, the refined poses and velocities are feedback to the network, generating loss function as:
\begin{equation}
    \mathbf{L}_{imu} = \mathbf{L}_{\Delta \mathrm{p}} + \mathbf{L}_{\Delta \mathrm{v}} + \mathbf{L}_{\Delta \mathrm{R}}
\end{equation}
\begin{equation}
    \begin{split}
        \mathbf{L}_{\Delta \mathrm{p}} = &\mathrm{R}_{w}^{b_{n-1}} \Big( {\mathbf{p}}_{b_{n}}^{w} - {\mathbf{p}}_{b_{n-1}}^{w} - {\mathbf{v}}_{b_{n-1}}^{w} \Delta t  \\& + \frac{1}{2} g^{w}\Delta t^2 \Big) - \hat{\bm{\alpha}}^{b_{n-1}}_{b_{n}}(\mathbf{\hat{b}}_a,\mathbf{\hat{b}}_g)
    \end{split}
\end{equation}
\begin{equation}
    \begin{split}
        \mathbf{L}_{\Delta \mathrm{v}} = &\mathrm{R}_{w}^{b_{n-1}} \Big( {\mathbf{v}}_{b_{n}}^{w} - {\mathbf{v}}_{b_{n-1}}^{w} +  g^{w}\Delta t \Big) \\& - \hat{\bm{\beta}}^{b_{n-1}}_{b_{n}}(\mathbf{\hat{b}}_a,\mathbf{\hat{b}}_g)
    \end{split}
\end{equation}
\begin{equation}
    \begin{split}
        \mathbf{L}_{\Delta \mathrm{R}} = &\mathrm{Log} \left( \mathbf{R}(\hat{\bm{\gamma}}_{b_{n}}^{b_{n-1}}(\mathbf{\hat{b}}_g))^\mathrm{T} \mathrm{R}_{w}^{b_{n-1}} \mathrm{R}_{b_{n}}^{w} \right)
    \end{split}
    \label{eq:rloss}
\end{equation}

The IMU loss is almost the same with preintegration residuals Eq.(\ref{eq:ppre})-(\ref{eq:rpre}), while the constrained entities shifted from the system's states to the parameters of the networks. Besides, there are some other slight differences to be noticed. 1) The loss function constrains the bias predicted by the network rather than the bias updated by VIBA. 2) In the feedback loop, we manually set the weights for each loss function (all set to 1 here), while in VIBA, the weights for each residual are determined by the covariance \cite{manifold}. 3) Bias residuals Eq.(\ref{eq:ba})-(\ref{eq:bg}) are not included in the loss function.

The feedback-based self-supervised update mechanism enhances the consistency between the learning and classic optimization modules, collectively improving the overall performance.

\subsection{Online continual learning and VIO system}
\label{subsec: online}
Performance degradation is often encountered when a learning algorithm is transferred to an unseen environment. In such cases, adaptive fine-tuning of the network is a common practice. However, in our scenario, network fine-tuning often requires customized data preprocessing and training strategy, posing an additional workload and delaying the deployment of the VIO system.

\begin{table*}[htbp]
    \renewcommand\arraystretch{1.1}
	\centering
	\begin{tabular}{l|ccccc|cccccc}
		\toprule
		      \multirow{2}*{Methods} & \multicolumn{5}{c|}{Online adaptation to EuRoC \cite{euroc}} & \multicolumn{6}{c}{Online adaptation to TUM-VI \cite{tumvi}} \\
        \cline{2-12}
            ~ & MH1 & MH2 & MH3 & MH4 & MH5 & room1 & room2 & room3 & room4 & room5 & room6 \\
		\midrule
            DPVO \cite{dpvo} & 0.087 & 0.055 & 0.158 & 0.137 & 0.114 & 0.251 & 0.503 & 0.261 & 0.085 & 0.197 & 0.059 \\
            \midrule
            $\mathbf{V}_{PT}+\mathbf{I}_{RW}$ & 0.093 & 0.070 & 0.154 & 0.141 & 0.137 & 0.271 & 0.487 & 0.258 & 0.073 & 0.245 & 0.068 \\
            \multirow{2}*{$\mathbf{V}_{AD}+\mathbf{I}_{RW}$} & 0.043 & 0.042 & 0.108 & 0.107 & 0.095 & 0.143 & 0.369 & 0.237 & 0.063 & 0.165 & 0.053 \\
            ~ & 54$\%\uparrow$ & 40$\%\uparrow$ & 30$\%\uparrow$ & 24$\%\uparrow$ & 31$\%\uparrow$ & 47$\%\uparrow$ & 24$\%\uparrow$ & 8$\%\uparrow$ & 14$\%\uparrow$ & 33$\%\uparrow$ & 22$\%\uparrow$ \\
            \multirow{2}*{$\mathbf{V}_{PT}+\mathbf{I}_{AD}$} & 0.085 & 0.065 & 0.150 & 0.130 & 0.136 & 0.249 & 0.460 & 0.251 & 0.067 & 0.242 & 0.059 \\
            ~ & 9$\%\uparrow$ & 7$\%\uparrow$ & 3$\%\uparrow$ & 8$\%\uparrow$ & 1$\%\uparrow$ & 8$\%\uparrow$ & 6$\%\uparrow$ & 3$\%\uparrow$ & 8$\%\uparrow$ & 1$\%\uparrow$ & 13$\%\uparrow$ \\
            \multirow{2}*{$\mathbf{V}_{AD}+\mathbf{I}_{AD}$} & 0.041 & 0.039 & 0.109 & 0.100 & 0.094 & 0.138 & 0.351 & 0.231 & 0.065 & 0.158 & 0.052 \\
            ~ & 56$\%\uparrow$ & 44$\%\uparrow$ & 29$\%\uparrow$ & 29$\%\uparrow$ & 31$\%\uparrow$ & 49$\%\uparrow$ & 28$\%\uparrow$ & 10$\%\uparrow$ & 11$\%\uparrow$ & 36$\%\uparrow$ & 24$\%\uparrow$ \\
		\bottomrule
	\end{tabular}
    % \vspace{7pt}	
    \caption{Evaluation of online adaptation ability of the proposed modules to new environments, reported with metric-scaled RMSE ATE. In this table, $PT$, $RW$ denotes the pre-trained visual model and IMU bias model of random walk, and $AD$ is the abbreviation of adaptation. We statistic the improvements of each adaptation module than our baseline model $\mathbf{V}_{PT}+\mathbf{I}_{RW}$ in percentage.}
    \label{tab:online}
\end{table*}

To reconcile the contradiction between VIO deployment and network adaptation, we propose a ``learning within VIO'' strategy known as online continual learning. In this mechanism, the VIO system can be considered an automatic dataloader, responsible for organizing and optimizing training data and feeding it to the neural network. In addition to being able to run continuously alongside VIO, our online continual learning mechanism is also highly flexible. 1) We can start or stop it at any time without interrupting the execution of VIO. 2) We can train the visual correspondence or the IMU bias predictor independently or simultaneously.

The VIO system can be summarized as follows:

\textbf{Initialization}: Our VIO system starts with an initialization process, which includes map initialization and IMU initialization. After IMU initialization, we align the body coordinate and recover the metric scale of the pose and the map. The initialization process provides the system with a good initial state and builds the platform for online learning.

\textbf{Tracking}: The pipeline of tracking are mainly shown in Fig.\ref{pipline}. We compute the initial pose of the incoming frame by IMU state propagation and add it to the factor graph. The factor graph maintains a sliding window containing states of the latest 10 keyframes. For efficiency considerations, during online continual learning, the visual constraints for each keyframe are only derived from several keyframes preceding and following it. Otherwise, the visual constraints of each keyframe may encompass several keyframes with optical flow magnitudes below a threshold, forming a co-visibility relationship.

\textbf{Feedback}: After tracking, we utilize the feedback-based self-supervised updates described in Sec.\ref{subsec: network} to implement online continual learning for the networks. During online adaptation, we fix the weights of the feature encoder and fine-tune the predictors for both visual and IMU. All inputs and constraints for the networks come from the factor graph, including the images, IMU measurements, and state estimates, generating loss functions as Eq.(\ref{eq:pholoss}-\ref{eq:rloss})

\textbf{Keyframing}: After each tracking session, keyframe culling is performed. Our keyframing strategy is similar to DPVO \cite{dpvo}. However, considering the temporal constraints of IMU, the gap between two keyframes should not exceed 3 frames.

\section{Experiments}
\label{sec: expreiments}

\subsection{Implementation details}
\label{subsec:mplementation}

Our method is implemented by Python and PyTorch \cite{pytorch}, with specific components like VIBA utilizing C++ and CUDA programming for acceleration. The visual network requires pre-training, while the IMU bias network does not. In the following content, we will present the methods of pre-training and online continual learning separately, primarily focusing on the latter.

\textbf{Datasets}:
The TartanAir\cite{tartanair} dataset is adopted to pre-train our visual model, which is a large-scale simulated dataset widely used in visual learning tasks.
 We choose the EuRoC \cite{euroc} and TUM-VI \cite{tumvi} datasets for online continual learning and validation. The EuRoC is captured by a UAV visual-inertial platform, while the TUM-VI is collected by a handheld visual-inertial device. Both datasets include environments with complex lighting conditions and intricate motion patterns, making them widely used in VIO. To give the quantified results, we align the estimated trajectories with the provided ground-truth and compute the Root Mean Square Error (RMSE) of the Absolute Trajectory Error (ATE) \cite{SLAMbenchmark}.

\begin{table*}[htbp]
	\centering
	\begin{tabular}{lc|cccccccccccc}
		\toprule
		      Sequence & Sensor & MH1 & MH2 & MH3 & MH4 & MH5 & V11 & V12 & V13 & V21 & V22 & Avg. \\
		\midrule
            MSCKF\cite{msckf2} & M+I & 0.42 & 0.45 & 0.23 & 0.37 & 0.48 & 0.34 & 0.20 & 0.67 & 0.10 & 0.16 & 0.34 \\
            OKVIS\cite{okvis} & M+I & 0.33 & 0.37 & 0.25 & 0.27 & 0.39 & 0.094 & 0.14 & 0.21 & 0.09 & 0.17 & 0.23 \\
            ROVIO\cite{rovio} & M+I & 0.21 & 0.25 & 0.25 & 0.49 & 0.52 & 0.10 & 0.10 & 0.14 & 0.12 & 0.14 & 0.23 \\
            VINS-Mono\cite{vins} & M+I & 0.15 & 0.15 & 0.22 & 0.32 & 0.30 & 0.079 & 0.11 & 0.18 & 0.08 & 0.10 & 0.17 \\
            Kimera\cite{kimera} & S+I & 0.11 & 0.10 & 0.16 & 0.24 & 0.35 & 0.05 & 0.08 & 0.07 & 0.08 & 0.10 & 0.13 \\
            Online VIO\cite{onlinevio} & M+I & 0.14 & 0.13 & 0.20 & 0.22 & 0.20 & 0.05 & 0.07 & 0.16 & 0.04 & 0.11 & 0.13 \\
            VI-DSO\cite{vidso} & M+I & 0.062 & \bf{0.044} & 0.117 & 0.132 & 0.121 & 0.059 & 0.067 & 0.096 & 0.040 & 0.062 & 0.08 \\
            DM-VIO\cite{dmvio} & M+I & 0.065 & \bf{0.044} & 0.097 & 0.102 & \bf{0.096} & 0.048 & \bf{0.045} & \bf{0.069} & \bf{0.029} & \bf{0.050} & \bf{0.06}\\
            % ORB-SLAM3\cite{orbslam3} & M+I & 0.062 & 0.037 & 0.046 & 0.075 & 0.057 & 0.049 & 0.015 & 0.037 & 0.042 & 0.021 & 0.04 \\
		\midrule
            iSLAM\cite{islam} & S+I & 0.500 & 0.391 & 0.656 & 1.285 & 1.088 & 0.521 & 0.405 & 0.397 & 0.421 & 0.580 & 0.58 \\
            SelfVIO\cite{selfvio} & M+I & 0.19 & 0.15 & 0.21 & 0.16 & 0.29 & 0.08 & 0.09 & 0.1 & 0.11 & 0.08 & 0.15 \\
            \textbf{Ours} & M+I & \bf{0.050} & 0.055 & \bf{0.069} & \bf{0.092} & 0.124 & \bf{0.035} & \bf{0.045} & 0.073 & 0.052 & 0.086 & 0.07 \\
		\bottomrule
	\end{tabular}
    % \vspace{7pt}	
    \caption{Evaluation of visual-inertial odometry systems on EuRoC dataset, with RMSE ATE ($m$), SE(3)-aligned. The upper list is all classic VIO methods; the bottom is learning-based systems. The letters ``M'', ``S'' and ``I'' denote monocular, stereo, and IMU.}
    \label{tab:euroc}
\end{table*}

\textbf{Pre-training settings}:
The visual part of our network (module (A)and(C) in Fig.\ref{pipline}) requires pre-training. Following the strategy in DPVO \cite{dpvo}, we train our visual model on the TartanAir dataset \cite{tartanair} for 240000 iterations with a batch size of 1. Note that due to minor modifications in our networks compared to DPVO \cite{dpvo}(discussed in Sec.\ref{subsec: network}), the performance after pre-training may not be identical. 

\textbf{Online continual learning settings}:
After pre-training the visual network, we perform online continual learning on the EuRoC and TUM-VI datasets. Our PC configuration includes an Intel i9-9900 CPU and an NVIDIA GTX 3090 GPU with 24GB of VRAM.

During online continual learning, our VIO performs tracking and carries out gradient backpropagation for each incoming frame. This process constitutes one iteration of training. For tracking stability, we may not update the networks at every iteration. Completing an entire sequence is considered as one epoch, and online continual learning requires training multiple epochs, involving the continuous replay of one sequence.

For the EuRoC dataset, we perform online continual learning in MH\_01. Similarly, we conduct continual learning in room1 of the TUM-VI dataset. In both settings, we replay the sequences for 60 epochs. The visual predictor's network weights update every 100 iterations, while the IMU bias predictor's weights update with each iteration. The learning rate for the visual predictor is set to be $1\times10^{-5}$. For the IMU bias predictor, since it is not pre-trained, we conduct visual BA for the first 30 epochs to ensure tracking stability, with the learning rate of $1\times10^{-4}$. After that, we perform VIBA with the learning rate of $1\times10^{-6}$.

\begin{figure}
  \centering
  \includegraphics[width=0.48\textwidth]{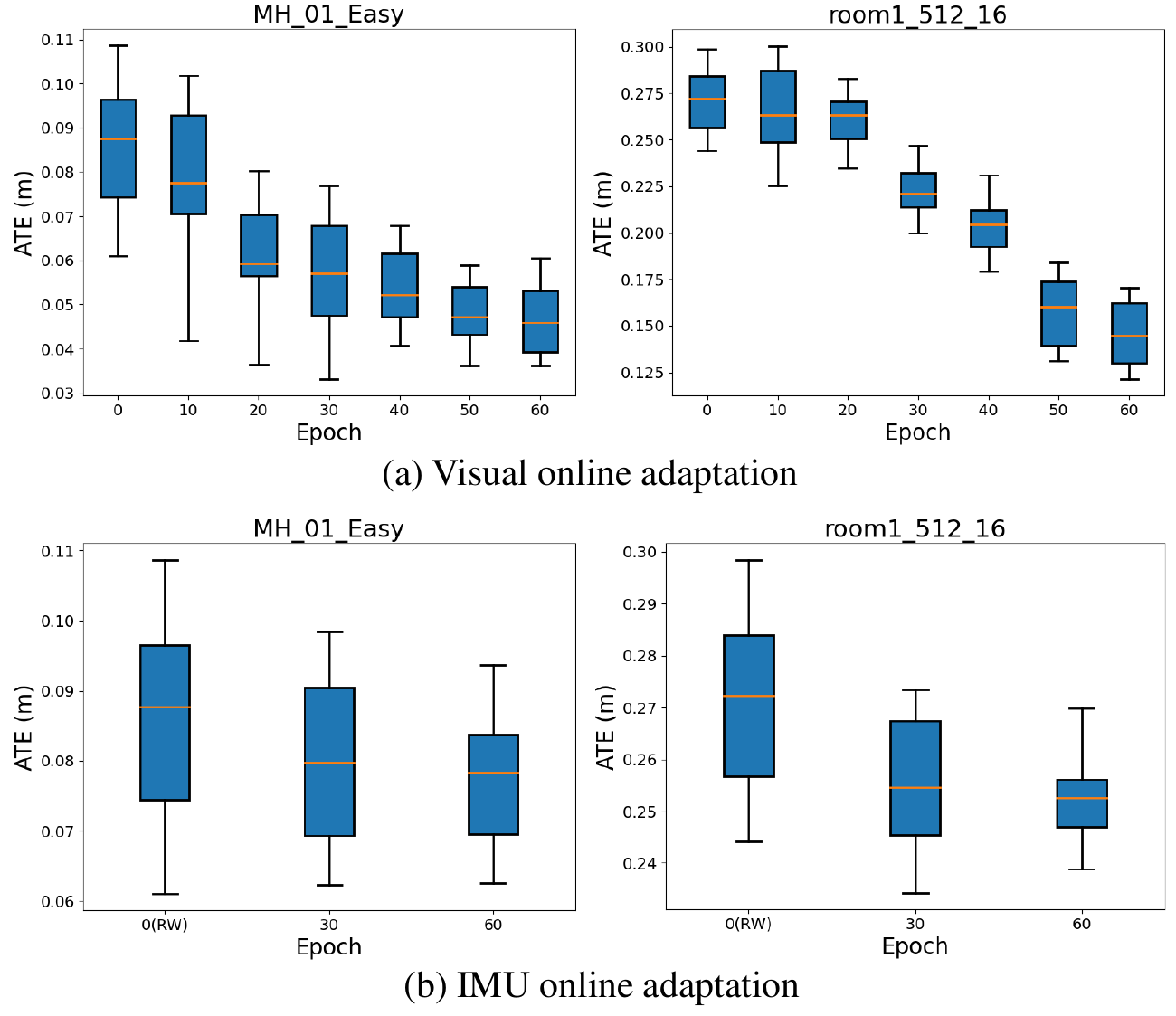}
  \caption{RMSE ATE changes in MH\_01 (EuRoC) and room1 (TUM-VI) during online continual learning of visual and IMU.}
  \label{fig:online performance}
\end{figure}

\subsection{Evaluation of online continual learning}
To evaluate the effectiveness of online continual learning for visual and IMU bias networks, we take the pre-trained visual model and IMU bias random walk as our baseline model and then compare it with visual adaptation, IMU adaptation, and joint adaptation. The results are summarized in Tab.\ref{tab:online}. To eliminate the influence of the metric scale, all results are Sim(3) aligned and averaged over five trials.

Additionally, since the visual networks of our method are mainly borrowed from DPVO \cite{dpvo}, it's reasonable to present their results for reference, as shown in the first row of Tab.\ref{tab:online}, where the results are Sim(3) aligned and are the median of five trials.

Compared to the baseline, online continual learning for either visual or IMU networks improves trajectory accuracy, and the joint adaptation achieves the best performance, resulting in over 10\% improvements across all sequences. 

To further validate the continual adaptability of our method, we perform additional statistical tests during online continual learning. We conduct 10 validation experiments for the visual model after every 10 epochs to assess its trajectory accuracy and distribution. The overall statistical results are illustrated in Fig.\ref{fig:online performance}(a). We can observe that the ATE distribution of trajectories continually decreases during adaptation. For the IMU model, although the continual learning of IMU bias contributes less significantly to the improvement of trajectory accuracy compared to the visual model, we statistically compare its errors with those of the random walk model in 10 experiments after every 30 epochs. As shown in Fig.\ref{fig:online performance}(b), the IMU bias model presents lower trajectory drift and variance than bias random walk, thus leading to more robust performance.

\begin{figure*}[htbp]
  \centering
  \includegraphics[width=1.0\textwidth]{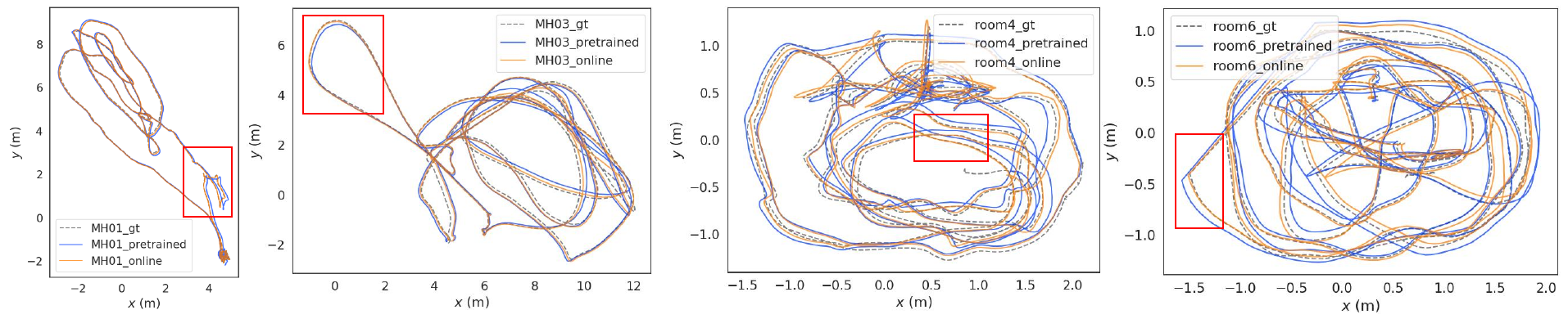}
  \caption{Estimated trajectories comparison on EuRoC (two sub-figures on the left) and TUM-VI dataset (two sub-figures on the right).}
  \label{fig:trajectory}
\end{figure*}

Two examples of each dataset are selected for trajectory comparison of our model, with and without online continual learning. As shown in Fig. \ref{fig:trajectory}, the trajectories of our online adaptation model are closer to the ground-truth than that of our pre-trained model on four sequences, as clearly indicated by the red boxes.

\begin{table}
	\centering
	\begin{tabular}{l|ccccc}
		\toprule
		      \makecell{Seq.} & \makecell{ROVIO \\ stereo} & \makecell{VINS \\ mono} & \makecell{OKVIS \\ stereo} & \makecell{DM-VIO \\ mono} & \makecell{\textbf{Ours} \\ mono} \\
		\midrule
            corr.1 & 0.47 & 0.63 & 0.33 & 0.19 & \bf{0.14} \\
            corr.2 & 0.75 & 0.95 & 0.47 & 0.47 & \bf{0.28} \\
            corr.3 & 0.85 & 1.56 & 0.57 & \bf{0.24} & 0.50 \\
        	corr.4 & \bf{0.13} & 0.25 & 0.26 & \bf{0.13} & 0.27 \\
        	corr.5 & 2.09 & 0.77 & 0.39 & \bf{0.16} & 0.21 \\
        	mag.1 & 4.52 & 2.19 & 3.49 & 2.35 & \bf{1.18} \\
            mag.2 & 13.43 & 3.11 & 2.73 & 2.24 & \bf{1.17} \\
            mag.3 & 14.80 & \bf{0.40} & 1.22 & 1.69 & 3.74\\
            mag.4 & 39.73 & 5.12 & \bf{0.77} & 1.02 & 2.49\\
            mag.5 & 3.47 & 0.85 & 1.62 & \bf{0.73} & 1.43\\
            mag.6 & X & 2.29 & 3.91 & \bf{1.19} & 3.12\\
            room1 & 0.16 & 0.07 & 0.06 & \bf{0.03} & 0.05\\
            room2 & 0.33 & 0.07 & 0.11 & 0.13 & \bf{0.04}\\
            room3 & 0.15 & 0.11 & 0.07 & 0.09 & \bf{0.02}\\
            room4 & 0.09 & 0.04 & \bf{0.03} & 0.04 & 0.04\\
            room5 & 0.12 & 0.20 & 0.07 & 0.06 & \bf{0.04}\\
            room6 & 0.05 & 0.08 & 0.04 & \bf{0.02} & 0.04\\
		\midrule
		      Avg. & 5.071 & 1.099 & 0.949 & \bf{0.634} & 0.867 \\
		\bottomrule
	\end{tabular}
    \caption{Evaluation of VIO methods on TUM-VI dataset, with RMSE ATE ($m$), SE(3)-aligned. The corr. and mag. represent corridor and magistrale sequences, respectively.}
    \label{tab:tum-vi}
\end{table}

\subsection{Evaluation of overall performance}

To evaluate the overall performance of our VIO system, we compare our method with state-of-the-art VIO approaches on the EuRoC \cite{euroc} and TUM-VI \cite{tumvi} datasets. Our system constructs a more extensive set of keyframe association constraints containing temporally adjacent keyframes and spatially neighboring keyframes, as described in Sec. \ref{subsec: online}. We perform online continual learning on MH\_01 of the EuRoC dataset and room1 of the TUM-VI dataset, then generalize to other sequences in the same dataset. The results are computed as the median of three trials. All trajectories are in metric scale and SE(3)-aligned with the ground-truth.

\textbf{Results on EuRoC:} As presented in Tab.\ref{tab:euroc}, we choose the state-of-the-art classic VIO methods and learning-based approaches as comparison. Compared with classic methods, our method outperforms most approaches in terms of RMSE ATE and is comparative to the performance of the DM-VIO \cite{dmvio}. Furthermore, our method exhibits significantly superior performance compared to other learning-based VIO methods. We also observe the performance of our method exceeds the VIO system Kimera \cite{kimera} and iSLAM \cite{islam}, both with stereo-inertial settings. 

\textbf{Results on TUM-VI:} To evaluate the generalization ability of our method, we also conduct experiments on another TUM-VI dataset \cite{tumvi}, which is a highly challenging handheld dataset with large-scale scenes, compared with the EuRoC dataset \cite{euroc}. 

We compare our method to the classic state-of-the-art VIO methods as presented in Lukas et al. \cite{dmvio}. The results are reported in Tab.\ref{tab:tum-vi}. Our method can achieve better accuracy on 7 sequences than alternative methods, even compared to the DM-VIO method, which shows the best result among all methods. However, in other sequences, DM-VIO reported better results than ours, which can be attributed mainly to its more robust initialization and long-term scale refinement.

\section{Conclusion}
\label{sec: conclusion}

This paper presents a novel VIO system named Adaptive VIO, which combines online continual learning with classic optimization. We employ neural networks to predict visual correspondence and IMU bias and then construct visual-inertial bundle adjustment to tightly couple both sensor measurements to refine the state estimation. The refined estimates can be fed back to the front-end networks that are updated through online continual learning, enabling our system to adapt to new environments. Experimental results illustrate that the online continual learning of our method can improve the overall system performance, whether for visual adaptation, IMU adaptation, or joint of them. Compared with classic and learning-based state-of-the-art VIO systems, our method can achieve competitive results and show potential adaptation to unseen scenarios. In the future, we plan to explore extending the online feedback mechanism to various networks and improving the robustness and efficiency of the system.

\section*{Acknowledgement}

We thank anonymous reviewers and AC for their fruitful comments and suggestions. This work is supported by the NSF, China (U22A2061, 62176010), and 230601GP0004.

{
    \small
    \bibliographystyle{ieeenat_fullname}
    \bibliography{ref}
}

% \appendix

\end{document}